\documentclass[10pt,twocolumn,letterpaper]{article}

\usepackage{cvpr}              

\usepackage{graphicx}
\usepackage{amsmath}
\usepackage{amssymb}
\usepackage{booktabs}

\usepackage{algorithm}
\usepackage[noend]{algpseudocode}
\usepackage{multicol}
\usepackage{multirow}
\usepackage{wrapfig}
\usepackage{enumitem}
\usepackage{array}
\usepackage{gensymb}
\usepackage{siunitx}
\usepackage{setspace}
\usepackage{pifont}

\newcolumntype{?}{!{\vrule width 1pt}}

\usepackage[pagebackref,breaklinks,colorlinks]{hyperref}

\usepackage[capitalize]{cleveref}
\crefname{section}{Sec.}{Secs.}
\Crefname{section}{Section}{Sections}
\Crefname{table}{Table}{Tables}
\crefname{table}{Tab.}{Tabs.}

\newcommand{\inter}{\textsc{Interactron}\xspace}

\def\confName{CVPR}
\def\confYear{2022}

\begin{document}

\title{Interactron: Embodied Adaptive Object Detection}

\author{Klemen Kotar and Roozbeh Mottaghi\\
 PRIOR @ Allen Institute for AI
}
\maketitle

\begin{abstract}
Over the years various methods have been proposed for the problem of object detection. Recently, we have witnessed great strides in this domain owing to the emergence of powerful deep neural networks. However, there are typically two main assumptions common among these approaches. First, the model is trained on a fixed training set and is evaluated on a pre-recorded test set. Second, the model is kept frozen after the training phase, so no further updates are performed after the training is finished. These two assumptions limit the applicability of these methods to real-world settings. In this paper, we propose Interactron, a method for adaptive object detection in an interactive setting, where the goal is to perform object detection in images observed by an embodied agent navigating in different environments. Our idea is to continue training during inference and adapt the model at test time without any explicit supervision via interacting with the environment. Our adaptive object detection model provides a 7.2 point improvement in AP (and 12.7 points in AP$_{50}$) over DETR \cite{carion2020end}, a recent, high-performance object detector. Moreover, we show that our object detection model adapts to environments with completely different appearance characteristics, and performs well in them. The code is available at: \url{https://github.com/allenai/interactron}.         
\end{abstract}

\begin{figure}[tp]
    \centering
    \includegraphics[width=\linewidth]{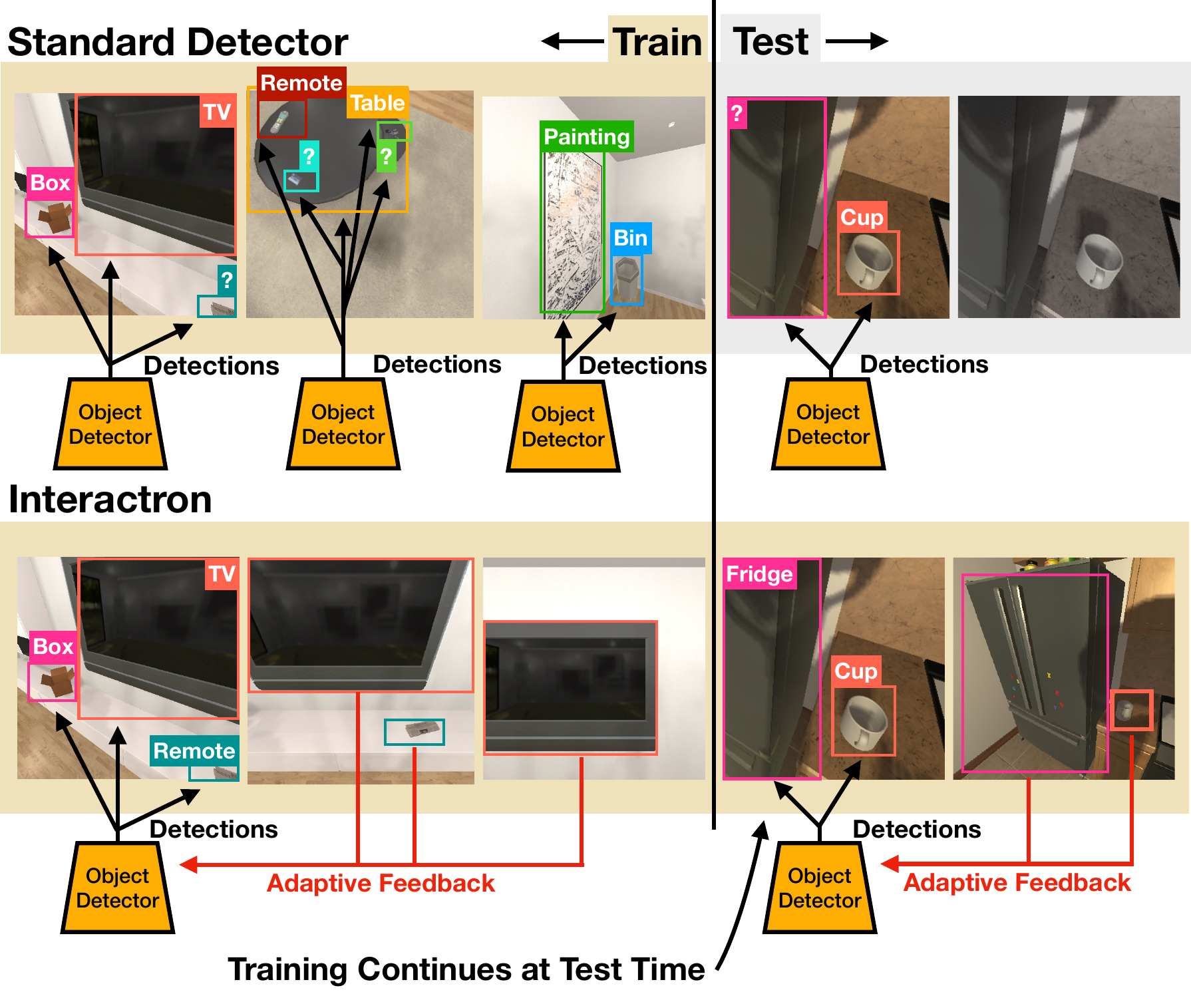}
    \caption{We introduce \inter\ a novel method for object detection. The idea is to adapt the detection model in an interactive setting during inference without any explicit supervision. The top row shows a standard detector that is kept frozen during inference. The bottom row shows our model that is updated during inference without any supervision by using future observations. }
    \vspace{-0.4cm}
    \label{fig:teaser}
\end{figure} 

\section{Introduction}

Object detection has been a central problem in computer vision since the inception of the field. There has been an extensive literature over the past decades proposing various methods ranging from constellation \cite{fischler1973representation,felzenszwalb2000efficient,dpm}, region-based \cite{Gu2009RecognitionUR,Sande2011SegmentationAS,Vijayanarasimhan2011EfficientRS}, and hierarchical \cite{jin2006context,zhu2008unsupervised,Schnitzspan2009DiscriminativeSL} models to the more recent powerful CNN \cite{girshick2014rich,girshick2015fast,ren2015faster} and Transformer \cite{carion2020end,zhu2020deformable,chen2021pix2seq} based models to tackle this problem. Typically, there are two main assumptions in these works: (1) There is a fixed training set and a test set. (2) The model is frozen after the training stage (i.e., it cannot be updated) and is evaluated on the pre-defined test set.

These assumptions pose certain limitations for object detection in real world applications. First, in many applications (e.g., autonomous driving or home assistant robots), the model continuously receives new observations from the environment. The new observations might help the model correct its belief. For example, a partially occluded object might not be detected confidently in the current frame, but there might be a better (unoccluded) view of that object in later observations. The model should use this signal to improve its confidence in similar situations in the future. Second, freezing the weights after training inhibits further improvement and adaptation of the model. We believe there are strong self-supervised signals in the inference phase that an embodied agent can leverage via interacting with its environment to adapt the model. There has been work to adapt object detector in an unsupervised way (e.g., \cite{su2020adapting,vs2021mega,wang2021domain,d2020one}). However, they assume a pre-recorded set of observations during inference. 

The idea of the proposed method is \emph{to continue training during inference} while \emph{interacting} with an environment. Our hypothesis is that interacting with the environment enables the embodied agent to capture better observations during inference leading to better adaptation and higher performance. In stark contrast to common object detection works, there is no distinct boundary between training and inference phases, and the model learns to take actions and adapt without any explicit supervision during inference. More specifically, there is an agent that interacts within indoor environments and relies on an object detector trained fully supervised to recognize objects. Our goal is to improve the object detection performance by adapting the model during inference while the agent interacts with the environment according to a learned policy (Figure~\ref{fig:teaser}). During training, the agent \emph{learns} a loss function using the supervised data, i.e. it learns to mimic the gradients produced during training using the labeled data. During inference, there is no supervision available for object detection. However, the model can generate gradients for the input images. Therefore, the model can be updated at inference time using the generated gradients. Basically, the model is updated without any explicit supervision at test time.

We evaluate our adaptive object detection model, referred to as Interactron\footnote{Inspired from Detectron \cite{Detectron2018}, the popular object detection framework.}, using the AI2-THOR \cite{ai2thor} framework which includes 125 different object categories that appear in 120 indoor scenes. The task is to detect objects in all frames that the agent observes while navigating in a scene. Our experiments show that by learning to adapt, the DETR \cite{carion2020end} model, which is a recent, high-performance object detection model, improves by 7.2 points in mAP. In addition to this strong result, we show that our adaptive model trained on AI2-THOR achieves near par results with a model trained with full supervision for the Habitat \cite{habitat} framework, which includes scenes with completely different appearance characteristics.

In summary, we propose an embodied adaptive object detection approach, where the model is updated during both training and inference. This approach is in contrast to traditional object detection, where the network is frozen after training. The model learns to adapt during inference via interaction with the environment and without any explicit supervision. We show our model significantly outperforms strong non-adaptive baselines and generalizes well to environments with different appearance distributions.

\section{Related Work}

\noindent \textbf{Object detection.} Various methods have been proposed to tackle the problem of object detection. Part-based and region-based models \cite{dpm,segdpm,wang2013regionlets} were among the high-performing methods before the emergence of CNN based approaches. CNN based detectors \cite{girshick2014rich,girshick2015fast,ren2015faster,redmon2016you,liu2016ssd} and the recently proposed Transformer based approaches \cite{carion2020end,zhu2020deformable,chen2021pix2seq} have achieved remarkable performance on detection benchmarks. One of the main assumptions of these works is that the model is kept frozen after training i.e. the weights of the model cannot change at test time. In contrast, in this paper, the model is updated in a self-supervised way to improve the detection performance. There have also been various works in segmentation and detection that adapt the network \cite{tsai2018learning,su2020adapting,vs2021mega,wang2021domain,d2020one}. However, they have access to only a fixed set of images and there is no mechanism to interact with an environment.

\noindent \textbf{Embodied adaptive methods.} Our detection model falls in the category of models that adapt at test time. We describe a few examples of these approaches for embodied tasks. \cite{nagabandi2018learning} propose a method to adapt online to novel terrains, crippled body parts, and highly-dynamic environments. \cite{li2020self} propose an algorithm to enable visual odometry networks to continuously adapt to new environments. \cite{shangda20} propose a domain adaptation method for visual navigation so methods trained in simulation better generalize to real environments. However, it is different from the adaptive methods in the sense that it has access to some target-domain images during training. \cite{wortsman2019learning} propose a meta-learning based approach to adapt to new test environments for the task of navigation towards objects. In \cite{li2020unsupervised}, there is a meta-learner that learns a set of transferable navigation skills. The agent can then quickly adapt to combine these skills when the navigation-specific reward is provided. \cite{wang2020visual} learn to adapt to new camera configurations using a few examples at test time for the task of vision and language navigation. \cite{lekkala2019meta} meta-learn a set of tasks (environment configurations) to better generalize to new tasks using a few samples. In contrast to these approaches, we focus on improving object detection. More importantly, unlike these approaches (except \cite{wortsman2019learning}), we propose to learn a loss function instead of relying on a pre-defined loss function.

Our method shares similarities with continual learning \cite{thrun1995lifelong} approaches. However, most continual learning works focus on passive non-embodied scenarios (e.g., \cite{rebuffi2017icarl,shmelkov2017incremental,camoriano2017incremental}). A recent work by \cite{liu2021lifelong} proposes a continual learning method for a navigation scenario. Continual learning works typically focus on learning without forgetting while our objective is to adapt efficiently to test scenarios without any supervision.  

\noindent \textbf{Active vision.} Active vision \cite{blake1993active} typically involves an agent that moves in an environment to have a better perception for the defined task or to perform the task more efficiently. The active vision literature addresses various types of tasks such as 3D reconstruction \cite{krainin2011autonomous,dunn2009next,dunn2009developing}, object recognition \cite{ammirato2017dataset,jayaraman2016look,johns2016pairwise}, 3D pose estimation \cite{doumanoglou2016recovering,zeng2018robotic,wu20153d}, and 3D scene modeling \cite{blaer2007data}. The main difference of our approach with these works is that our model is updated on-the-fly based on a loss that is \emph{learned} self-supervised (as opposed to the typical manually defined measures of uncertainty). \cite{yang2019embodied} is closer to our work and infers a policy to better recognize objects. Again, it differs from our approach since they freeze their model after training and it is based on full supervision. \cite{Nilsson2021EmbodiedVA} propose to actively select a view in a scene and request annotation for that view. In contrast, our approach is adaptive and does not request annotations. \cite{fang2020better} use pseudo-labels for self-supervised training of object detection. In contrast, we learn a policy and more importantly, continue training during inference. 

\noindent \textbf{Embodied self-supervision.} There are various works that learn self-supervised representations via embodied interactions \cite{pot2018self,Lohmann2020LearningAO,mitash2017self,PirkOnlineObjects2019,eitel2019self}. Our goal is different in that we learn a loss function in a self-supervised fashion to change the weights on an object detector in a new environment to adapt to that environment. \cite{chaplot2021seal} addresses self-supervised learning for object detection by relying on a pose sensor, depth images, and 3D consistency. In contrast, we use only RGB images and do not rely on a perfect pose assumption.
\section{Embodied Adaptive Learning}
\label{sec:emb_adp_learn}
In this section, we introduce our approach to applying embodied, adaptive learning during inference to the object detection task. The main idea is that we do not freeze the weights of the model after training and instead let the model adapt during inference without any explicit supervision while an embodied agent explores the environment. 

\subsection{Task Definition}
We first introduce a new flavor of the object detection task, suited for interactive environments (such as AI2-THOR \cite{ai2thor} or Habitat \cite{habitat}). The task consists of predicting a bounding box and category label for every object in the egocentric RGB frame of an embodied agent. Formally we are given a scene $S \in \mathbf{S}$, and a position $p$ and asked to predict every object $o \in \mathbf{O}_{S,p}$, the set of all objects visible in $f_{S, p}$ (the egocentric RGB frame at position $p$ in scene $S$). The agent is also allowed to take $n$ actions from the action set $\mathbf{A}$ according to some policy $\mathcal{P}$ and record the $n$ additional RGB frames that it observed. We call the sequence of the $n$ frames observed by our agent $\mathbf{F}$. We then use some model $\mathcal{M}$, which takes $\mathbf{F}$ as input, to predict a bounding box and class label for every object in $\mathbf{O}_{S,p}$ (for a certain vocabulary of object categories). Note that we perform the detection only on the initial frame. Otherwise, the agent will be encouraged to ``cheat" by simply moving to an area with few easily detectable objects. 

For every position $p$ in every scene $S$, there are many possible sequences of frames $\mathbf{F}$ as there are many trajectories that the agent can explore. We call these sequences of frames rollouts, and we define the set of all rollouts for a given scene $S$ and position $p$ as $\mathbf{R}_{S, p}$.
Finally, since there are many scenes and positions, each of which can be the starting point for many rollouts, we can define the set $\mathbf{R_{all}}$ of all the possible rollouts for all the possible positions in all of our scenes, such that $\mathbf{F} \in \mathbf{R}_{S, p} \subset \mathbf{R_{all}}$. 
In summary, each instance of an interactive object detection task $\mathcal{T}$ contains a scene $S$ and a starting position $p$ and is drawn from some distribution $d(\mathcal{T}, \mathbf{S})$ of all task instances given a set of scenes.

\subsection{Standard Approaches}
\label{sec:standard-app}
 The most trivial approach to solving this problem is by using an off-the-shelf detector $\mathcal{M}_{exist}$ and simply performing object detection on the initial frame $f_{S, p}$. Here our policy $P_{no-op}$ is simply to not take any actions at all. We can boost the performance by pre-training the object detector on data from the same domain as our interactive environment.

 A more powerful approach will use a random policy to move the agent and collect several frames around the starting position $p$. Then a multi-frame model $\mathcal{M}_{mf}$ can be trained to perform object detection on the initial frame using all of the frames as input. A model trained on such sequences can learn to leverage the multiple perspectives of the objects collected by the agent as it moves around to improve object detection.

\subsection{Adaptive Learning}
\label{sec:adapt}
Intuitively, training an object detector in one particular local area of an environment (be it a room, building, or scene) increases the performance of the object detector on other nearby frames in that local area, since these environments (and in fact the natural world) are continuous. We confirm this intuition by empirical results, so we proceed to formulate this task as a meta learning problem where each instance of the interactive object detection task $\mathcal{T}$ represents a new task to fit to. At training time this abstraction works well as we can treat each frame in $\mathbf{F}$, and their corresponding ground truth labels, as a task example and apply a version of the MAML algorithm~\cite{finn2017model}. We can then produce an object detector $\mathcal{M}_{meta}^{\theta}$, parameterized by $\theta$. We train this model by doing a forward pass with all the frames in $\mathbf{F}$, then computing the backward pass by using the ground truth labels and an object detection loss $\mathcal{L}_{det}$. We then take a gradient step and update our parameters such that $\theta' = \theta - \alpha \nabla_{\theta} \mathcal{L}_{det}(\theta, \mathbf{F})$. We then optimize the model by minimizing the detection loss $\mathcal{L}_{det}$. We repeat this process on many tasks from $d(\mathcal{T}, \mathbf{S}_{train}$), where $\mathbf{S}_{train}$ is a set of training scenes.

At test time, however, this approach is infeasible, as we are not given labels for any of the frames. We can overcome this by adding another loss, one that is not based on the labels but rather just the frames in $\mathbf{F}$. This loss can be hand designed, or it can be learned. Taking inspirations from~\cite{Yu2018OneShotIF,wortsman2019learning}, we learn the loss function. In our case, we use a learned loss produced by a model called the adaptive loss or $\mathcal{L}_{ada}^\phi$ parameterized by $\phi$ that takes as input all of the frames in $\mathbf{F}$ as well as the predictions $\mathcal{M}_{meta}^{\theta}(\mathbf{F})$  to produce a gradient used for dynamic adaptation. There is no explicit objective for the learned loss. Instead, we simply encourage that minimizing this loss improves the detection ability of our model.
Thus the learning objective for this model is
\begin{equation}
    \underset{\theta, \phi}min \sum_{\mathbf{F} \in \mathbf{R_{all}}} \mathcal{L}_{det}(\theta - \alpha \nabla_{\theta} \mathcal{L}_{ada}^{\phi}(\theta, \mathbf{F}), \mathbf{F})
\label{eq:mainobj}
\end{equation}

As mentioned above $\mathcal{L}_{det}$ is not available at test time, so the parameters of $\mathcal{L}_{ada}^{\phi}$ are frozen and only $\mathcal{M}_{meta}^{\theta}$ is trained according to $\mathcal{L}_{ada}^{\phi}$.
This method allows us to dynamically adapt our object detector to its local environment, using the information contained in the frames in $\mathbf{F}$, which were obtained by a random policy $P_{rand}$.

\subsection{Interactive Adaptive Learning}

In standard adaptive and meta learning applications, we generally operate under the assumption that the distribution of data samples for each task is fixed and can not be influenced by us. In the interactive setting this is not true, as the samples we use for adaptation are collected by our agent. Formally, at each time step our agent takes an action $a$ according to some policy $\mathcal{P}$\footnote{We denote learned policies by $\mathcal{P}$ and pre-defined ones with $P$.}, which takes as input all of the previous frames it has seen. Following the intuition that not all samples provide the same quality and quantity of information, we can learn a policy $\mathcal{P}_{int}$, which is a neural network parameterized by $\rho$ and optimize it to guide the agent along a sequence of frames $\mathbf{F}$ that will allow $\mathcal{M}_{meta}$ to easily adapt to the new task.

In order to learn $\mathcal{P}_{int}$, we must first find a way to assign a value to each rollout $\mathbf{F}$ obtained by our actions. We do this by measuring the similarity of the gradients of $\theta$ produced by the detection loss $\mathcal{L}_{det}$ (computed based on the labeled data) and those produced by the learned loss $\mathcal{L}_{ada}$. More specifically, we measure the $\ell_1$ distance between the gradients produced by $\mathcal{L}_{det}$ using the ground truth labels of the first frame and the gradients produced by $\mathcal{L}_{ada}$ using the sequence of frames our agent collected. This way we are encouraging the agent to collect frames that will help the learned loss emulate the supervision provided by the ground truth labels. We call this value the Initial Frame Gradient Alignment $IFGA$ and define it for any sequence $\mathbf{F}$ as follows:

\begin{equation}
    IFGA(\mathbf{F}) = \sum \lvert \nabla_{\theta} \mathcal{L}_{ada}^{\phi}(\theta, \mathbf{F}) - \nabla_{\theta} \mathcal{L}_{det}(\theta, [f_{S,p}]) \rvert
\end{equation}

This allows us to extract another useful training signal from our learned loss. Note that we only compute the $IFGA$ of complete sequences of length $n+1$ (the initial frame plus the $n$ frames collected by the agent) as our learned loss estimator takes $n+1$ frames as input to compute the adaptive gradient. We can then define a full exploitation policy $P_{exp}$ which given any incomplete sequence of frames $\mathbf{F}_{inc}$ where $len(\mathbf{F}_{inc}) < n$ explores every possible completion of the sequence and outputs the action that leads toward a complete sequence of frames with the lowest $IFGA$. This ideal policy is computed during training by exploring every possible trajectory the agent can take from the starting frame $f_{S, p}$, but at test time this is not possible. Instead, we use a neural network $\mathcal{P}^{\rho}_{int}$ parameterized by $\rho$ as our policy and train it to clone the behavior of $P_{exp}$ using the following loss function:

\begin{equation}
    \mathcal{L}_{pol}(\mathcal{P}^{\rho}_{int}, \mathbf{F}) = - P_{exp}(\mathbf{F})^T \log \mathcal{P}^{\rho}_{int}(\mathbf{F})
\end{equation}

For a task where our agent is permitted to take $n$ steps in the environment we define the adaptive gradient step learning rate as $\alpha$, and the detector, learned loss, and policy learning rates as $\beta_1, \beta_2$ and $\beta_3$, respectively. Then we write down our interactive adaptive training algorithm as in Algorithm~\ref{alg:training}. The inference procedure for adaptive interactive learning is the same as the one for adaptive learning described in Section~\ref{sec:adapt}, with the exception that frames are not rolled out randomly by $P_{rand}$, but are rather obtained by following $\mathcal{P}_{int}$.

\label{algorithm} 

\begin{algorithm}[tp]
\caption{Training $(d(\mathcal{T}, \mathbf{S}_\text{train}), \theta, \phi, \rho, \alpha, \beta_1, \beta_2, \beta_3, n)$}
\label{alg:train}
\begin{algorithmic}[1]
\While{not converge}
    \For{mini-batch of tasks $\tau_i \in d(\mathcal{T}, \mathbf{S}_\text{train})$}
        \State{$\theta_i \gets \theta$}
        \State{$t \gets 0$}
        \State{$\mathbf{F}_i \gets [f_{S_i, p_i}]$}
        \While{$t < n$}
            \State{Sample action $a$ from $\mathcal{P}_{int}^{\rho}(\mathbf{F}_i)$}
            \State{Take action $a$ and collected frame $f$}
            \State{$\mathbf{F}_i \gets \mathbf{F}_i + [f]$}
            \State{$t \gets t + 1$}
        \EndWhile
        \State{$\theta_i \gets \theta_i - \alpha \nabla_{\theta_i} \mathcal{L}_{ada}^{\phi}(\theta_i, \mathbf{F}_i)$}
    \EndFor
    \State{$\theta \gets \theta - \beta_1 \sum_{i} \nabla_{\theta} \mathcal{L}_{det}(\theta_i, f_{S_i, p_i})$}
    \State{$\phi \gets \phi - \beta_2 \sum_{i} \nabla_{\phi} \mathcal{L}_{det}(\theta_i, \mathbf{F}_i)$}
    \State{$\rho \gets \rho - \beta_3 \sum_{i} \nabla_{\rho} \mathcal{L}_{pol}(\theta_i, \mathcal{P}^{\rho}_{int}(\mathbf{F}_i)) $}
\EndWhile
\end{algorithmic}
\label{alg:training}
\end{algorithm}

With this method, we are not only exploiting the ability of our model to fit to a local area of the scene, but also the fact that frames that would make good training examples for our model to be fit, also tend to contain useful information.

\section{Models}
\label{sec:models}
\begin{figure*}[tp]
    \centering
    \includegraphics[width=40pc]{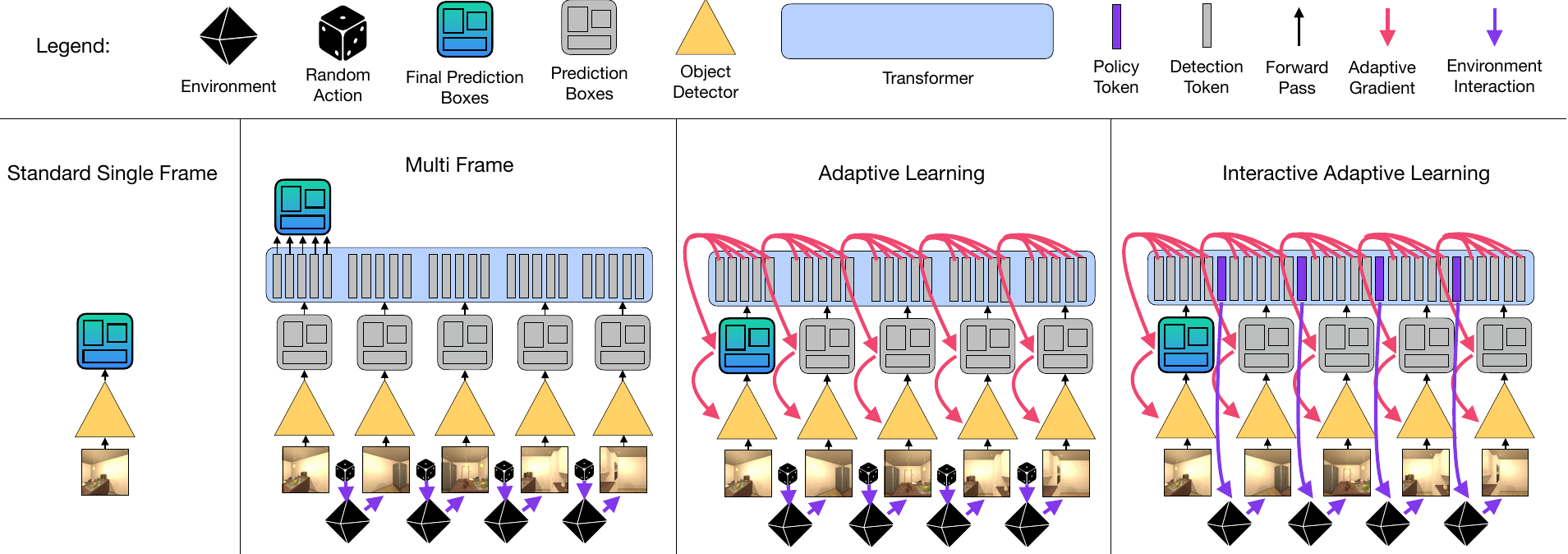}
    \vspace{-0.3cm}
    \caption{The architecture of the four main models presented in Section~\ref{sec:models}. A single frame baseline, which is just an off-the-shelf object detector, a multi-frame baseline which includes an inter-frame fusion Transformer and two \inter\ models (w/ and w/o a learned policy), which use a Transformer to learn a self-supervised loss function.}
    \label{fig:arch}
\end{figure*}

While the methods described in Section~\ref{sec:emb_adp_learn} are fundamentally model agnostic, certain architectures naturally fit this approach. In this section, we describe the specifics of the models studied in this paper. Figure~\ref{fig:arch} shows an overview of these models.

\subsection{\textsc{INTERACTRON} Model}
Our pipeline consists of two models: the Detector which performs the task of object detection and is adapted to the local environment at test time and the Supervisor which consists of the learned loss $\mathcal{L}_{ada}$ and learned policy $\mathcal{P}_{int}$ and is frozen at test time.

\textbf{The Detector} can be any off-the-shelf object detection model, but we use DEtection TRansformer (DETR)~\cite{carion2020end} for our experiments. It is architecturally simple and yet very powerful. It utilizes a ResNet backbone which produces image features and a Transformer model which attends to all of the features to produce object detection embeddings. Each object detection embedding is then passed through an MLP to extract bounding box coordinates and a predicted object class. Although we use Transformers elsewhere in our model, using a Transformer-based object detector is not a requirement of this architecture. (FasterRCNN~\cite{ren2015faster} results have also been provided in the Appendix).

\textbf{The Supervisor} is a Transformer model that functions as both the policy and the learned loss. Image features and object detection embeddings produced by the detector, collectively called \emph{detection tokens}, are passed into the Transformer. The transformer outputs of these tokens are passed through an MLP then reduced to a scalar to compute the adaptive gradient. In addition to these, a learnable \emph{policy token} is passed into the Transformer for each action the agent needs to take, and its output is used to compute the policy.

\textbf{Learned Loss Training} is performed by passing the outputs of the Detection Tokens through an MLP and taking the $\ell_2$ norm of all the features to obtain a scalar which is the adaptive learning objective.  The $\ell_2$ norm is a trick (also in \cite{Yu2018OneShotIF}) to combine the sequence of vectors produced by the Supervisor transformer into a single scalar loss. As the learning objective in Eq.~\ref{eq:mainobj} illustrates, the parameters of the Supervisor $\phi$ are optimized such that the gradients produced by the Supervisor $\nabla \theta \mathcal{L}_{ada}^{\phi}$ result in a reduction of the loss of the Detector. During training, the ground truth loss is computed from the object annotations, and the predictions made by the adapted Detector. The gradients used to adapt the Detector parameters are produced by the Supervisor. We can backpropagate through the adaptive gradients to update the Supervisor parameters, such that it produces better adaptive gradients (meta-training). At test time there is no object annotations, but the Supervisor has now been optimized to produce good gradients using the other frames and detections in the sequence.

\textbf{Policy Training} is performed by treating the problem as a sequence prediction task. We learn $n$ different embeddings to produce $n$ different Policy Tokens. The Transformer outputs of the Policy Tokens are passed through an MLP to produce the action probability distribution. All possible trajectories of length $n$ are explored during training and the policy is optimized to select the actions which lead to a complete rollout $\mathbf{F}$ with the lowest $IFGA$. When $n$ is small it is possible to roll out every trajectory, but as $n$ gets larger stochastic exploration or reinforcement learning methods are preferable, we leave this to future work. The Detection Tokens are fed into the Transformer one frame at a time, followed by a Policy Token, the output of which is used to predict the next action the agent should take. The model has access to all of the previous frames when deciding which action to take next. The adaptive gradient is only computed once all $n$ steps have been taken. Figure~\ref{fig:detarch} in the Appendix provides more details about our model. 

\begin{table*}[tp]
 \small  
  \centering
  \begin{tabular}{l?ccc?ccc?cc}         
  \toprule
                          & AP    & AP$_{50}$ & AP$_{75}$ & AP$_{S}$ & AP$_{M}$ & AP$_{L}$ & Adaptive & Policy\\\midrule
DETR~\cite{carion2020end}& 0.256   & 0.448  & 0.207  & 0.112   & 0.368  & 0.605  &    & No Move \\
Multi-frame              & 0.288   & 0.517  & 0.299  & 0.132   & 0.452  & 0.697  &    & Random \\
\inter-Rand (ours)       & 0.313   & 0.551  & 0.320  & 0.167   & 0.477  & 0.710  &  \ding{51}  & Random \\
\inter\ (ours)           & \textbf{0.328}   & \textbf{0.575}  & \textbf{0.331}  & \textbf{0.174}   & \textbf{0.480}  & \textbf{0.733}  &  \ding{51}  & Learned Policy \\
  \bottomrule
  \end{tabular}
  \vspace{-0.3cm}
  \caption{\textbf{Object detection results.} We compare our method \inter\ with a set of baseline approaches. The standard COCO detection metrics have been reported. Our method provides a massive gain compared to the single-frame DETR~\cite{carion2020end} baseline. }
  \vspace{-0.3cm}
  \label{tab:main_result}
\end{table*}
\subsection{Ablation Models}
\noindent\textbf{The \inter-Rand Model} is essentially the same as the interactive adaptive learning model described above, except that it does not feed policy tokens to the Transformer, and instead utilizes a random policy $P_{rand}$.

\noindent\textbf{The Multi-Frame Baseline} is architecturally the same as the Adaptive Learning Model, but instead of using the Transformer as a learned loss function $\mathcal{L}_{ada}$, it uses it as a fusion layer, combining the detector output of all the frames in the sequence to produce detections for the first frame. This architecture corresponds to the model $\mathcal{M}_{mf}$ described in Section~\ref{sec:standard-app}.

\noindent\textbf{The Single Frame Baseline} is simply an off-the-self detector (in our case DETR) that has been pre-trained on some data including images from our interactive environment and corresponds to $\mathcal{M}_{exist}$.

\section{Experiments}

We perform an extensive set of experiments to show the advantages of our \inter\ model.\footnote{The results in this section are different from those in the CVPR 2022 version. The trends are still the same and we observe a huge gap between the performance of the proposed approach and the baselines. The discrepancy is due to a bug in our evaluation code that selected the best checkpoint on the test set.} We compare our method with a variety of baselines: a non-adaptive baseline, a non-adaptive baseline that aggregates information across multiple frames, and an adaptive baseline that explores a scene randomly. We also perform an experiment on an environment different from the one used for training and evaluate how well our model adapts. We also perform a set of ablation experiments to better analyze the proposed model.

\textbf{Implementation details.} We conduct the bulk of our experiments in the AI2-iTHOR~\cite{ai2thor} interactive environment, as it offers many similarly sized scenes, fast rendering, and the ability to perform domain randomization. For our main task, we consider 5 frames (the initial frame plus 4 frames collected by the agent interacting with the environment). We use the action set $\{$MoveForward, MoveBackward, RotateLeft, RotateRight$\}$ and $30$ degree rotation angles.  We perform our evaluations with 300x300 images. Performing the object detections at higher resolutions could offer a significant performance improvement, but it also comes with significantly increased computational complexity so we leave exploring different resolutions for future work.

For our main experiment, we train \inter\ models using the train set $d(\mathcal{T}, \mathbf{S}_{train})$ and test it on $d(\mathcal{T}, \mathbf{S}_{test})$. We train our model for 2,000 epochs on the training set, using SGD to perform the meta training step and the Adam optimizer to train the learned loss and policy models. For training details see Appendix~\ref{app:training}. We ensure that all possible trajectories are explored during the entire training run. The single frame baseline is just the pre-trained object detector, while the multi-frame baseline is trained on $d(\mathcal{T}, \mathbf{S}_{train})$ for 1,000 epochs using the Adam optimizer. We use the standard COCO metrics for results.

\textbf{Dataset.} We collect two datasets, $d(\mathcal{T}, \mathbf{S}_{train})$ and $d(\mathcal{T}, \mathbf{S}_{test})$, where $\mathbf{S}_{train}$ consists of AI2-iTHOR~\cite{ai2thor} training and validations scenes (100 scenes in total) and ${S}_{test}$ consists of AI2-iTHOR test scenes (20 scenes). The datasets consist of scene id $S$, agent starting positions $p$ as well as the labels for every object visible from the starting position $\mathbf{O}_{S,p}$. The starting positions (which consist of positional and rotational coordinates) are randomly sampled from all available positions for a given scene. After each sampling, the locations of all the objects in a given scene are randomized. We uniformly draw a total of 1000 pairs $S, p$ from the training and validation scenes and 100 pairs from the test scenes. Note that the agent can explore different trajectories depending on the policy. So only the initial frames in the test set are fixed. 

We also collected a pre-training dataset of 10K frames from the training scenes with object detection annotations to train the base object detector. We employ the same sampling methods as above. We create a new set of detection classes which is the union of all the object categories from LVIS~\cite{gupta2019lvis} and AI2-iTHOR~\cite{ai2thor}. This results in a total of 1,235 object categories and a ``background" category. For our AI2-iTHOR evaluations, we only use the 125 iTHOR object classes, ignoring the others.

\textbf{Pre-training the model.} We pre-train the model, using the DETR~\cite{carion2020end} codebase on a dataset of 124K LVIS images and 10K images from the AI2-iTHOR pre-training dataset. We use the standard 500 epoch training schedule.

\subsection{\inter Results}

 Table~\ref{tab:main_result} shows that our method outperforms the baselines by a significant margin. We compare our method with a non-adaptive single-frame baseline DETR~\cite{carion2020end}, a non-adaptive multi-frame baseline, referred to as ``Multi-frame", an adaptive baseline with a random policy, referred to as ``\inter-Rand".

Our random policy \inter\ uses the same data and neural network architecture as the multi-frame baseline, yet it outperforms it, showcasing the merits of adaptive training even when our agent is not following an optimal policy. Our full \inter\  model further widens the performance gap by selecting good frames for computing the learned loss. The overall improvement between the off-the-shelf detector (DETR~\cite{carion2020end}) that the single frame baseline represents and our model is 7.2 AP (and 12.7 AP$_{50}$).

\subsection{Transfer Results}
One of the key goals of this work is adapting our model to novel environments. In order to evaluate the cross-environment adaptability performance of \inter\ models, we evaluate them on the Habitat~\cite{habitat} environment that has completely different appearance characteristics (natural images vs synthetic AI2-iTHOR images). We train our model in the AI2-iTHOR environment and perform adaptive inference in the Habitat environment. We test our model on a dataset of Habitat task instances, which are generated similarly to our AI2-iTHOR task instances (see Appendix~\ref{app:data}). For this set of tasks we only used the intersection of the iTHOR categories which we trained our model on, and the Habitat categories (22 object categories in total). Table~\ref{tab:transfer_result} shows \inter\ models trained on AI2-iTHOR  images are able to perform almost on par with a detector pre-trained with full supervision in Habitat. This shows that our method allows a model to fit to a new environment without access to training data or labels from that environment as well as a model that does have access to the labeled data in that environment. It is interesting to note that for small objects, \inter\ outperforms the baseline approaches (refer to AP$_S$ column). This indicates that our method may in fact leverage the signal in the later frames to update its belief about small or partially visible objects which are typically hard to detect.

\begin{table}
  \scriptsize
  \centering
  \tabcolsep=0.04cm
  \begin{tabular}{l|c|c|c|c|c|c|}         
  \toprule
                          & Dataset                      &     AP   &   AP$_{50}$   &   AP$_{S}$ & AP$_{M}$   & AP$_{L}$\\\midrule
DETR~\cite{carion2020end}&  LVIS + AI2-iTHOR             &  0.17    &   0.20   &   0.08     &  0.17   &   0.30   \\
DETR~\cite{carion2020end}&  LVIS + AI2-iTHOR + Habitat   &  \textbf{0.24}    &   \textbf{0.30}   &   0.13     &  \textbf{0.22}   &   \textbf{0.39}   \\          
\inter-Rand              &  LVIS + AI2-iTHOR             &  0.22    &   0.26   &   0.14     &  0.19   &   0.32   \\        
\inter\                  &  LVIS + AI2-iTHOR             &  0.22    &   0.27   &   \textbf{0.16}    &  0.20   &   0.32   \\
  \bottomrule
  \end{tabular}
  \vspace{-0.3cm}
  \caption{\textbf{Transfer results.} We train our model on the AI2-iTHOR~\cite{ai2thor} framework and perform adaptive inference in the Habitat~\cite{habitat} environment. }
  \vspace{-0.2cm}
  \label{tab:transfer_result}
\end{table}

\subsection{Ablations}

\textbf{No Training at test.} We profile the performance contribution of our learned loss $\mathcal{L}_{ada}$ by measuring the performance of our model on the test set without it. We roll out the trajectory according to $\mathcal{P}_{int}$, then simply omit applying the gradient update according to $\mathcal{L}_{ada}$. Table~\ref{tab:train_test} shows the meta trained model, without the train-at-test gradient update performs significantly worse than our full model, showcasing the contribution of training at test time. In fact, without the test time adaptation, our model performs worse than the off-the-shelf detector baseline.

\begin{table}
  \centering
  \small
  \begin{tabular}{l|c|c}         
  \toprule
                         & Train at test                &     AP$_{50}$  \\\midrule
\inter\                  &      yes      &     \textbf{0.58}            \\
\inter\                  &      no       &     0.28            \\
\inter-Rand              &      yes      &     0.55            \\
Multi-frame              &      no       &     0.52            \\
DETR~\cite{carion2020end}&      no       &     0.45            \\
  \bottomrule
  \end{tabular}
   \vspace{-0.3cm}
  \caption{\textbf{Ablation - No train at test.}}
  \vspace{-0.3cm}  
  \label{tab:train_test}
\end{table}

\textbf{Varying number of frames.} Since we have shown that gathering information from 5 frames is more useful than just relying on a single frame, it naturally follows to inquire if adding even more frames helps improve our model even further. To test this, we train \inter\ models with rollouts of length 7 and 9, respectively. We slow down the training schedule and increase the number of epochs when training these, as there are significantly more possible trajectories to explore with these sequence lengths (see Appendix~\ref{app:training} for details). Table~\ref{tab:diffno} shows the results. We find that any improvement gained by adding more frames is within the training noise of our model.

\begin{table}
  \centering
  \small
  \begin{tabular}{l|c|c|c}         
  \toprule
  No. of frames                & 5     & 7     & 9          \\\midrule\midrule
  \inter AP$_{50}$             & 0.575 & 0.568 & 0.579     \\
  Multi-frame AP$_{50}$        & 0.517 & 0.518 & 0.521     \\
  \bottomrule
  \end{tabular}
  \vspace{-0.3cm}  
  \caption{\textbf{Ablation - Varying number of frames.}}
  \vspace{-0.2cm}
  \label{tab:diffno}
\end{table}

\textbf{Multiple copies of the same frame.}
We explore training a model which sees five repetitions of the first frame instead of five unique frames. Training a policy with this model is meaningless, so we use just the \inter-Rand model. Table~\ref{tab:sameframe} show the results. The model trained to look at just the first frame performs better than the off-the-shelf DETR model, but worse than the model looking at 5 different frames. The improvement can be attributed to the extra learning capacity of the supervisor Transformer model. This verifies that our adaptive approach in fact benefits from extracting information from all of the frames in our trajectory to adapt the model to the current environment.

\begin{table}
  \centering
  \small
  \begin{tabular}{l|c}         
  \toprule
                                     &     AP$_{50}$         \\\midrule
\inter-Rand                          &     \textbf{0.55}            \\
\inter-Rand (Repeated First Frame)   &     0.51            \\
Multi-frame                          &     0.52            \\
DETR~\cite{carion2020end}            &     0.45            \\
  \bottomrule
  \end{tabular}
  \vspace{-0.3cm}  
  \caption{\textbf{Ablation - Repetition of the same frame.}}
    \vspace{-0.3cm}  
  \label{tab:sameframe}
\end{table}

\begin{figure*}[tp]
    \centering
    \vspace{-0.5cm}
    \includegraphics[width=40pc]{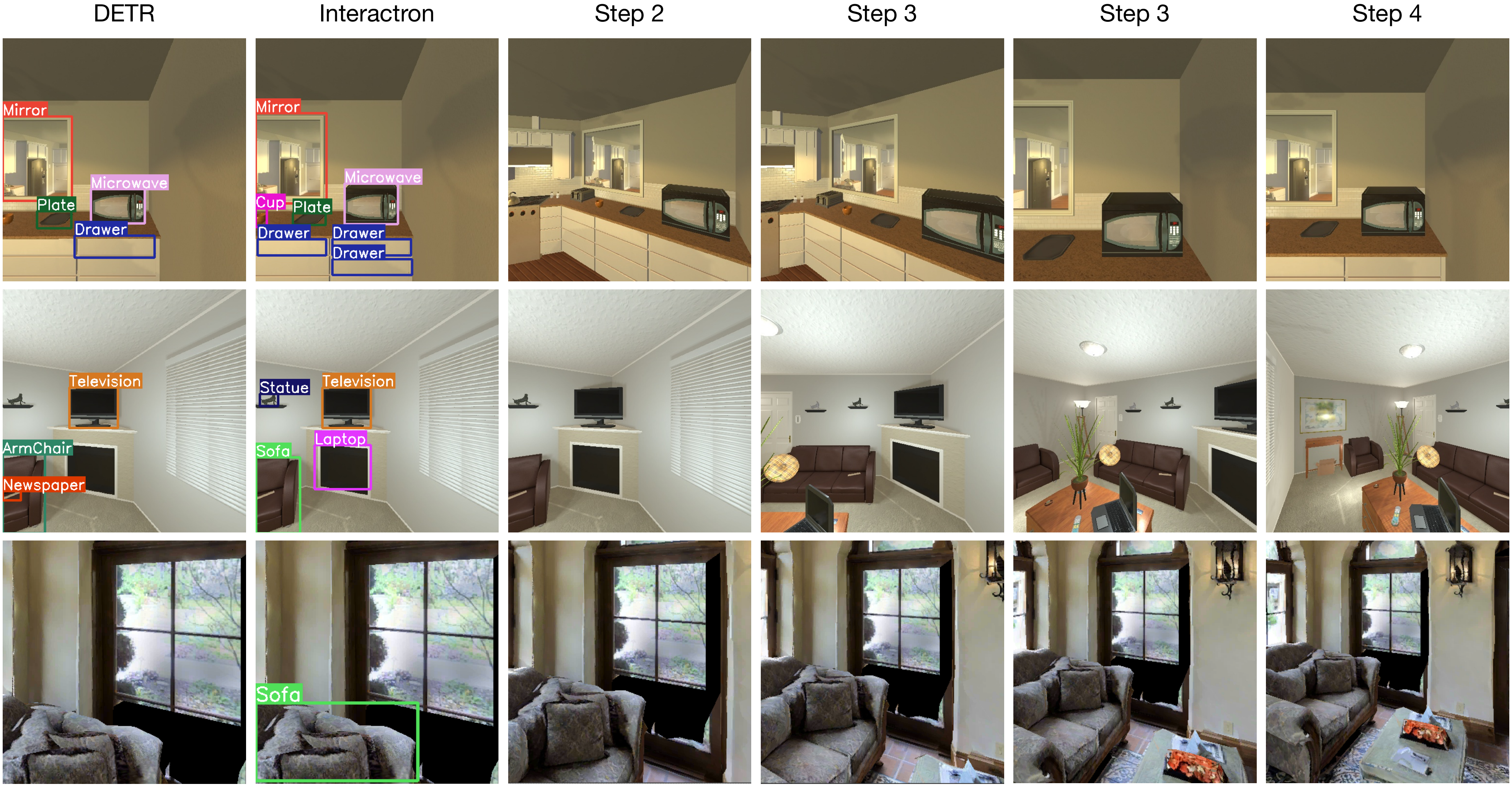}
    \vspace{-0.3cm}
    \caption{\textbf{Qualitative results.} in the AI2-iTHOR and Habitat environments. The first column displays the results produced by DETR~\cite{carion2020end}, the 2nd column displays the results of \inter, and the subsequent columns depict the interactive steps that the model took. Note that for the result shown in the third row, \inter\ has never seen Habitat images during training.}
    \vspace{-0.3cm}
    \label{fig:qual}
\end{figure*} 

\textbf{Variance analysis.} We repeat the AI2-iTHOR training to compute the performance variance between runs. We find that the standard deviation between the $AP_{50}$ of four different training runs to be 0.01.

\subsection{Qualitative Results}
Figure~\ref{fig:qual} shows a selection of rollouts produced by \inter. The bounding boxes displayed represent detections with a confidence score greater than 0.5. 

The first row showcases an example of situations in which \inter excels. The \emph{cup}, which is placed at the edge of the view of the agent is not detected by DETR, but \inter adjusts the position of the agent to get a better view of the \emph{cup}. Similarly, other \emph{drawers} are also detected by \inter.

The second row displays both a success and a failure case for our model. DETR only sees the arm of the \emph{sofa} and classifies it as an \emph{armchair}, while by looking from other angles, \inter discovers it is a \emph{sofa}. It is also able to correctly detect the \emph{statue}. On the other hand, DETR can detect the newspaper, which our model misses, and \inter falsely labels the fireplace as a laptop, an object it sees later in the rollout, and incorrectly associates with the position of the fireplace in the first image.

The third row shows an example of our transfer results in the Habitat environment. The agent starts positioned close to the sofa, so it is difficult to identify. DETR does not detect it while \inter takes a few steps backwards and is then able to correctly label it as a \emph{sofa}. Note that \inter\ has never seen Habitat images during training.

\vspace{-0.2cm}
\section{Discussion}
\vspace{-0.2cm}
\textbf{Constant Adaptation.}
One of the benefits of using an adaptive model versus one that simply views more frames at a time is that after the adaptation we are left with a model that only needs one frame to work well in the local area.

\textbf{Sub-policy.} If a certain complex task (such as navigation or object manipulation) requires a high confidence detection of the objects visible from a certain position, the interactive policy we propose can be used as a sub-policy.

\vspace{-0.1cm}
\subsection{Limitations}
\vspace{-0.2cm}
One of the main limitations of this approach is that it works only for a specific set of object categories (125 categories in our experiments), and it is not capable of learning about new categories. 
Another limitation is that the proposed policy training approach only works with short trajectories, which can be explored with reasonable space and time complexity. Further work can address alternative policy training approaches for tasks that require longer rollouts.
\vspace{-0.3cm}
\section{Conclusion}
\vspace{-0.2cm}
We introduce \inter\ an adaptive object detection model that adapts to its test environment without explicit supervision. The model gathers information about the scene via a learned policy deployed on an embodied agent that navigates in the environment. We show our approach substantially improves a state-of-the-art object detector (7.2 point improvement in object detection AP). Moreover, we showcase the strengths of our approach in adapting the model to new environments, performing near on par with a model trained fully supervised for that environment.

\noindent {\textbf{Acknowledgements.} We thank Aaron Walsman for the helpful discussions and Winson Han for his help with the figures. }

{\small
\bibliographystyle{ieee_fullname}
\bibliography{egbib}
}

\newpage
\appendix
\section*{Appendix}


\section{Pseudo Labels}
We investigate the viability of using pseudo-labels instead of a learned loss using the Multi-frame model and a simple pseudo-label method. We take the pre-trained Multi-frame model, and during evaluation apply a gradient update using the detections as labels, weighed by the detection confidence. This provides a small performance improvement of about 2 points with an AP$_{50}$ of 0.538. While our learned loss still outperforms this model, this ablation shows that a pseudo-label approach could also be viable in our problem setting.

\section{Model Details}
We use the DETR~\cite{carion2020end} model with a ResNet50 backbone for our detector. We freeze the backbone and Transformer encoder and only meta-train the decoder portion of DETR. We use the loss function based on the Hungarian matching algorithm described in the DETR paper as our ground truth detection loss $\mathcal{L}_{det}$. We set a maximum of 50 detections per image frame.

For our supervisor we use a 4 layer, 8 head Transformer based on the GPT architecture with an internal dimension of 512. Our supervisor also consists of two separate embedding layers for the Image Features and Object Detection Features respectively. What we refer to as ``Object Detection Features" are described as the output embeddings of the query tokens in the DETR paper. A positional embedding is learned for each token in the sequence. The supervisor also contains two decoders that consists of three consecutive linear layers each, with a hidden dimension of 512, separated by the GeLU non-linearity. One of the decoders consumes the outputs of the detection tokens and is used to compute the learned loss, while the other is used to decode the output of the Policy Tokens and produce a policy.

The learned loss is computed by taking the $\ell_2$ norm of the outputs of all the Detection Tokens passed through the decoder.

\begin{figure*}[tp]
    \centering
    \includegraphics[width=40pc]{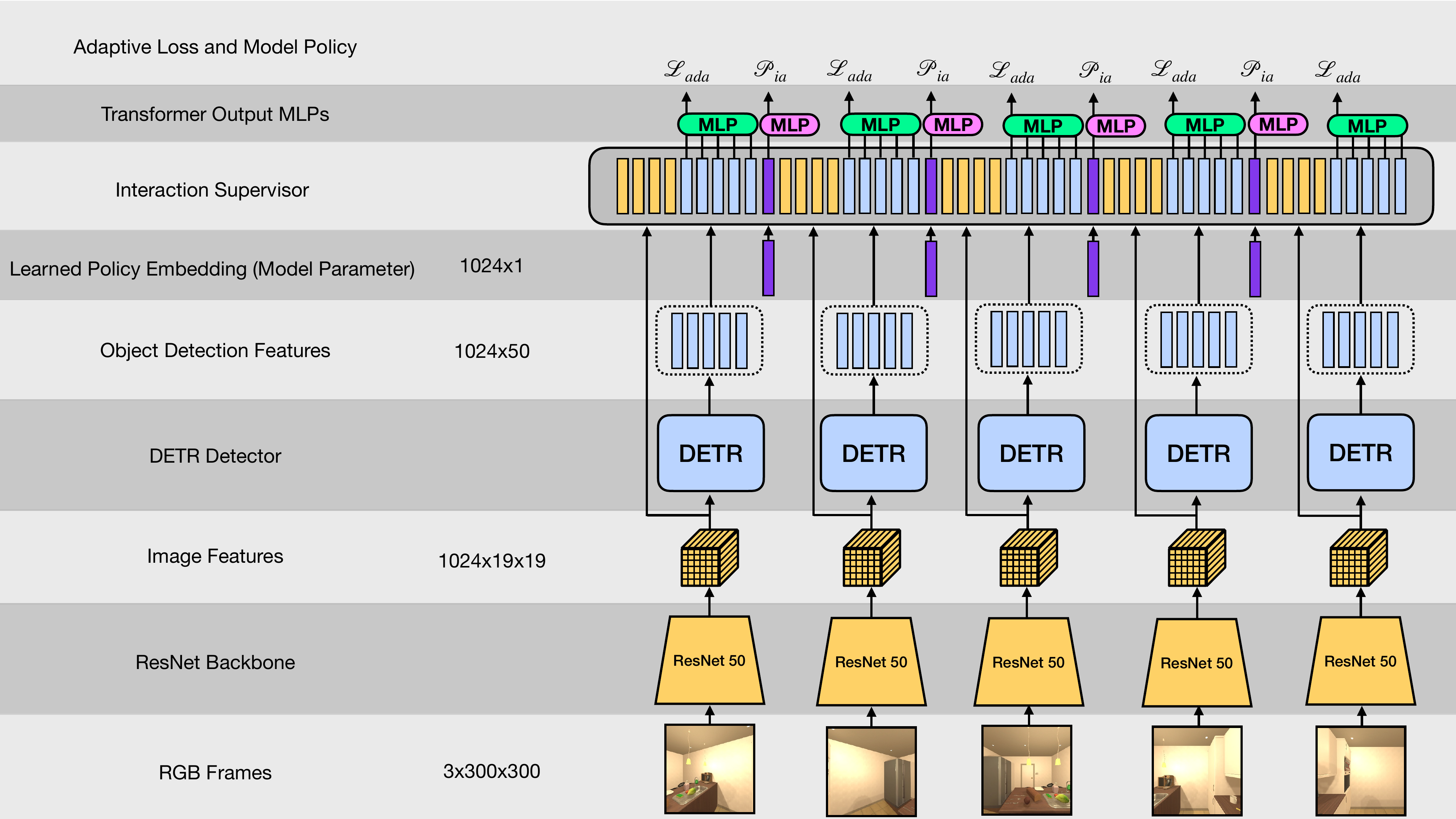}
    \caption{The detailed architecture of the \inter model. The first column describes each layer, while the second column provides the dimensions of the intermediate tensors where applicable.}
    \label{fig:detarch}
\end{figure*}

Figure~\ref{fig:detarch} illustrates a diagram of the \inter pipeline with tensor dimensions.

\section{Data Collection Details}
\label{app:data}
\subsection{iTHOR Data Collection Details}
For all of the iTHOR datasets the objects are labeled as one of the 1,235 classes which comprise the union of all the LVIS object categories and all the iTHOR object categories.

\textbf{The iTHOR Pre-Training Dataset} is collected from the 100 scenes in the train and val splits (scenes 0-25, 200-225, 300-325, 400-425). We uniformly sample frames from these scenes to collect a total of 10,000 frames and their annotations. A random shuffle of the placement of objects in the scene is performed before each sample is collected. If a frame has less than 3 objects visible in it, it is rejected and a new sample is drawn. The annotations are stored in LVIS format. The data is added to the LVIS dataset to collectively form the iTHOR+LVIS pre-training dataset.

\textbf{The iTHOR Training Set} is collected from the 100 scenes in the train and val splits. We uniformly sample starting positions from these scenes to collect a total of 1,000 starting location frames. A random shuffle of the placement of objects in the scene is performed before each sample is collected. If a frame has less than 3 objects visible in it, it is rejected and a new sample is drawn. In addition to this, every frame in every possible trajectory the agent could take from any sampled starting location is also collected. This implementation detail allows us to pre-cache every possible trajectory the agent could take to improve training efficiency.

\textbf{The iTHOR Test Set} is collected from the 20 scenes in the test split (scenes 25-30, 225-230, 325-330, 425-430). We uniformly sample starting positions from these scenes to collect a total of 100 starting location frames. If a frame has less than 3 objects visible in it, it is rejected and a new sample is drawn. In addition to this, every frame in every possible trajectory the agent could take from any sampled starting location is also collected. This implementation detail allows us to pre-cache every possible trajectory the agent could take to improve training efficiency.

\subsection{Habitat Data Collection Details}
For all of the Habitat datasets the objects are labeled as one of the 1,255 classes which comprise the union of all the LVIS object categories, all the iTHOR object categories and all the Habitat object categories.

\textbf{The Habitat Pre-Training Dataset} is collected from the 56 scenes in the MP3D train split. We uniformly sample frames from these scenes to collect a total of 10,000 frames and their annotations. Each scene has an equal number of samples, regardless of its size. If a frame has less than 3 objects visible in it, it is rejected and a new sample is drawn. The annotations are stored in LVIS format. The data is added to the iTHOR+LVIS pre-training dataset to collectively form the Habitat+iTHOR+LVIS dataset.

\textbf{The Habitat Test Dataset} is collected from the 11 scenes in the MP3D val split. We uniformly sample starting positions from these scenes to collect a total of 100 starting location frames. Each scene has an equal number of samples, regardless of its size. If a frame has less than 3 objects visible in it, it is rejected and a new sample is drawn. In addition to this, every frame in every possible trajectory the agent could take from any sampled starting location is also collected. This implementation detail allows us to pre-cache every possible trajectory the agent could take to improve training efficiency.

\section{DETR Pre-Training Details}
We pre-train the DETR object detector using the official DETR codebase, modified to accept the format of our dataset. We use the same training parameters as proposed in the original DETR paper to train one detector on the iTHOR+LVIS pre-training dataset and another on the Habitat+iTHOR+LVIS dataset. We scale all of the pre-training images such that one side has a dimension of 300.

\section{Training Details}
\label{app:training}
\subsection{Interactron Training}
The \inter model is trained for 2,000 epochs with the Adam optimizer, using an initial learning rate of $1\mathrm{e}{-4}$ and a linearly annealing schedule. Both the supervisor model producing the learned loss and policy and the detector itself are trained using these parameters. To stabilize the training we record the weights after each of the last 500 epochs and average them to produce the final model. The training takes approximately 120 hours using a single Nvidia RTX 3090.

During training we do not follow the policy of the model, but rather explore every possible trajectory the agent could take, from a given starting location. We then record the $IFGA$ value for each possible trajectory and update it whenever we revisit the same trajectory. We optimize our policy to always pursue the trajectory with the lowest recorded $IFGA$.

In slight contradiction to common nomenclature, one epoch does not represent a pass through every single frame in our dataset, as the agent can explore multiple trajectories from each starting location. For our 5 frame (4 step) experiments, for example, we can only guarantee that every possible frame in the dataset has been explored after 256 epochs of training. 

\subsection{Interactron 7 Frame Training}
For training the 7 frame model (where the agent takes 6 steps) we utilize a training regime of 5,000 epochs using the same initial learning rate and annealing schedule. This way we ensure that every possible trajectory the agent could take from each initial frame is explored multiple times.

\subsection{Interactron 9 Frame Training}
For training the 9 frame model (where the agent takes 8 steps) we utilize a training regime of 9,000 epochs using the same initial learning rate and annealing schedule. This way we ensure that every possible trajectory the agent could take from each initial frame is explored at least once.

\end{document}